\documentclass[sigconf]{acmart}
\AtBeginDocument{%
  }

\setcopyright{acmlicensed}
\copyrightyear{2025}
\acmYear{2025}
\acmDOI{XXXXXXX.XXXXXXX}
\acmConference[UIST '25]{ACM Symposium on User Interface Software and Technology}{September 28--October 1, 2025}{Busan, Korea}
\acmISBN{978-1-4503-XXXX-X/2018/06}
\usepackage{xcolor}

\usepackage{acmart-taps}

\usepackage{soul}
\sethlcolor{yellow} 
\usepackage{multirow}
\begin{document}

\title{Log2Plan: An Adaptive GUI Automation Framework Integrated with Task Mining Approach}

\author{Seoyoung Lee}
\email{leesy3891@sookmyung.ac.kr}
\orcid{0009-0002-0424-4785}
\affiliation{%
  \institution{Sookmyung Women's University}
  \city{Seoul}
  \country{Republic of Korea}
}

\author{Seobin Yoon}
\authornote{Both authors contributed equally to this research.}
\email{binsong2@sookmyumg.ac.kr}
\orcid{0009-0000-0113-8608}
\affiliation{
  \institution{Sookmyung Women's University}
  \city{Seoul}
  \country{Republic of Korea}
}

\author{Seongbeen Lee}
\authornotemark[1]  
\email{seongbeen@sookmyung.ac.kr}
\orcid{0009-0008-8704-3172}
\affiliation{
  \institution{Sookmyung Women's University}
  \city{Seoul}
  \country{Republic of Korea}
}
\author{Hyesoo Kim}
\email{sm2110883@sookmyung.ac.kr}
\orcid{0009-0002-1999-4060}
\affiliation{
  \institution{Sookmyung Women's University}
  \city{Seoul}
  \country{Republic of Korea}
}

\author{Joo Yong Sim}
\authornote{Corresponding author.}
\email{jysim@sookmyung.ac.kr}
\orcid{0000-0003-3779-7589}
\affiliation{
  \institution{Sookmyung Women's University}
  \city{Seoul}
  \country{Republic of Korea}
}

\renewcommand{\shortauthors}{Lee et al.}

\begin{abstract}
GUI task automation streamlines repetitive tasks, but existing LLM or VLM-based planner-executor agents suffer from brittle generalization, high latency, and limited long-horizon coherence. Their reliance on single-shot reasoning or static plans makes them fragile under UI changes or complex tasks. Log2Plan addresses these limitations by combining a structured two-level planning framework with a task mining approach over user behavior logs, enabling robust and adaptable GUI automation. Log2Plan constructs high-level plans by mapping user commands to a structured task dictionary, enabling consistent and generalizable automation. To support personalization and reuse, it employs a task mining approach from user behavior logs that identifies user-specific patterns. These high-level plans are then grounded into low-level action sequences by interpreting real-time GUI context, ensuring robust execution across varying interfaces. We evaluated Log2Plan on 200 real-world tasks, demonstrating significant improvements in task success rate and execution time. Notably, it maintains over 60.0\% success rate even on long-horizon task sequences, highlighting its robustness in complex, multi-step workflows.
\end{abstract}


\begin{teaserfigure}
  \includegraphics[width=\textwidth]{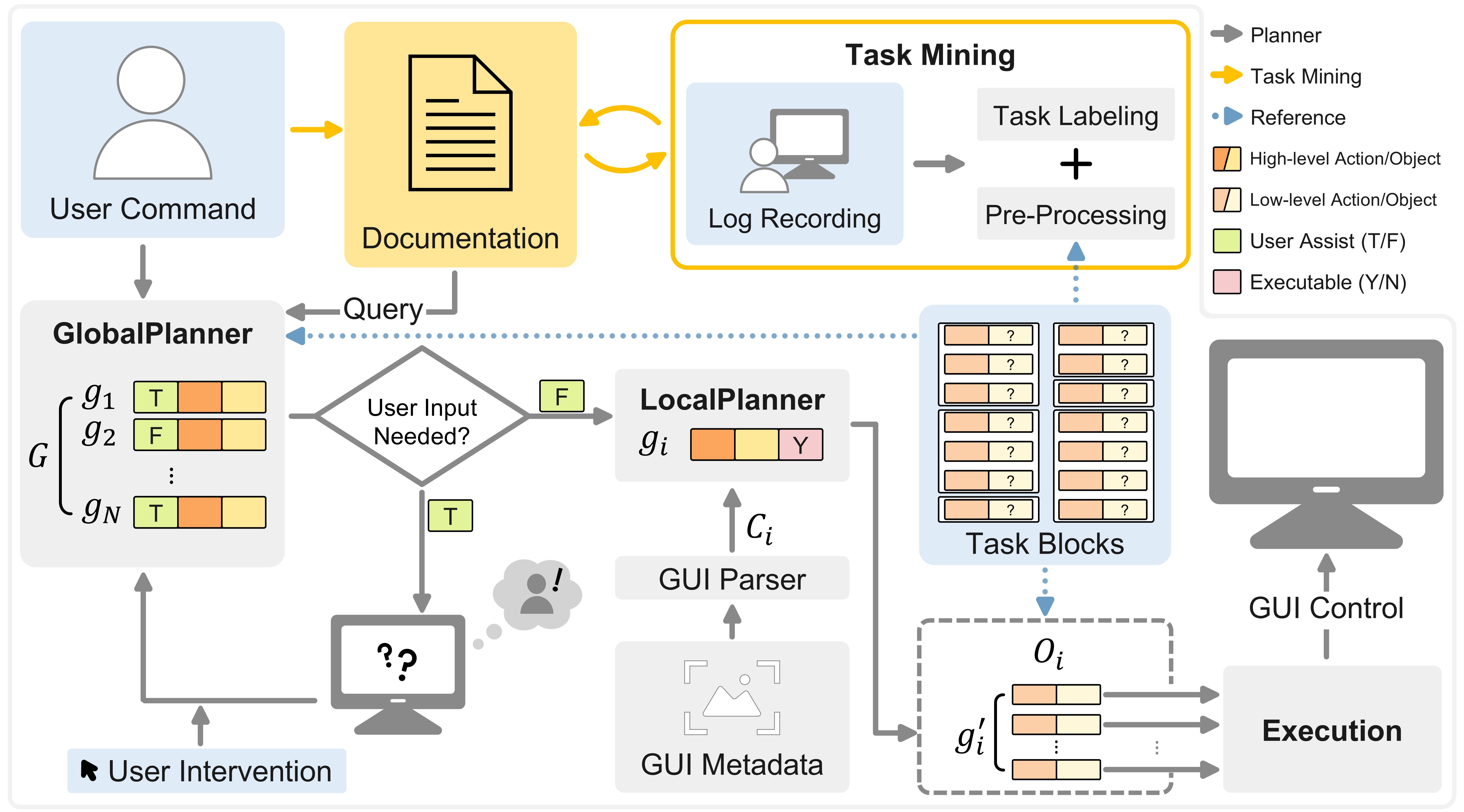}
  \caption{Overview of the proposed method : The Agent identify and label Tasks from collected log data. The user command is decomposed into a sequence of event-level tasks. Then, the LocalPlanner generates a task block-level plan, which corresponds to the event, optimized for the current GUI environment. Subsequently, the Execution phase carries out the low-level control.}
  \label{fig:main}
\end{teaserfigure}


\maketitle

\section{Introduction}
Large Language Models (LLMs) now demonstrate such strong capabilities in natural--language understanding and generation that they are being adopted as general purpose agents for personal assistance, software automation, and GUI testing\cite{raffel2020exploring,nguyen2024gui}. In particular, desktop based agents that can correctly interpret user commands and act inside real software environments have become a key research frontier\cite{gao2023assistgui,zhang2024ufo,qin2025ui,amazon2025novaact}.

Existing desktop agent approaches fall into three families: \emph{(i) Prompt based zero/ few shot agents} exploit pretrained models directly, yielding quick and flexible responses but suffering brittle behavior on multi step tasks\cite{yao2023react,yang2023mm,shinn2024reflexion}; \emph{(ii) Task specific fine tuned models} deliver high accuracy within narrow domains yet demand costly retraining and scale poorly to unseen tasks\cite{li2020mapping,hong2024cogagent,li2024ferret}; \emph{(iii) Planner--executor agents} couple an LLM planner with external tools, handling complex workflows at the price of engineering overhead and context window explosion\cite{zhang2024ufo,gao2023assistgui,qin2025ui}.

Despite their differences, all three approaches share fundamental limitations in real world GUI environments. They rely on static planners or rigid pipelines, making them fragile when layouts change or personalized reasoning is required. While task mining and RPA research highlight the value of recurring structures in user logs\cite{choi2022enabling,krosnick2022parammacros,huang2024automatic,riva2021etna}, existing agents ignore this opportunity. Replaying-based methods further show that small visual changes cause high failure rates\cite{deka2017rico,riva2021etna}. Moreover, vision encoding at each step inflates latency, while long plans overflow the model’s context window\cite{hong2024cogagent}.

Log2Plan addresses these gaps with a planning centric LLM framework that bridges symbolic task mining and retrieval augmented generation. It leverages user logs to learn and reuse generalizable task structures\cite{redis2024skill,yuan2024tasklama}. Concretely, Log2Plan (i) segments raw GUI logs into semantically meaningful \emph{action–object} units, and (ii) mines frequently repeated patterns to synthesize modular automation flows. Compared with record and replay baselines, this yields markedly better generalization and interpret-ability. At runtime, Log2Plan dynamically retrieves and recombines prior workflows to generate robust plans for new commands, enabling fast adaptation without model fine-tuning. By separating high-level intent from low-level execution, the LLM operates over structured, behavior-grounded representations, while a lightweight executor handles GUI interactions.

Our contributions are three fold:
\begin{enumerate}
    \item We formularize GUI interactions as \emph{action–object} pairs, enabling consistent and interpretable planning from both natural language commands and interaction logs.
    \item We identify recurring user behaviors in interaction logs and compose them into modular automation units that can be flexibly reused across diverse tasks.
    \item We propose a two level architecture that cleanly separates user intent from execution, allowing LLMs to plan in a robust and scalable manner.
\end{enumerate}  
Together, these contributions advance practical, user-centric, and broadly applicable LLM-powered desktop agents.

\section{Related Works}
\subsection{LLM/VLM-Based High-Level Planning and Execution}
Large language models (LLMs) and visual–language models (VLMs) have shifted GUI automation from rule-based scripts to agents that map natural language commands to interface actions. Early systems like SUGILITE~\cite{li2017sugilite} and APPINITE~\cite{li2018appinite} grounded user intents using structured UI metadata, while later approaches mined macros from user traces, though these remained brittle to layout changes~\cite{pan2022automatically,riva2021etna,krosnick2022parammacros}.

With advances in multimodal encoders, models began processing raw screenshots. Spotlight~\cite{li2022spotlight} narrows attention to widget candidates, and AssistGUI~\cite{gao2023assistgui} combines GPT-Vision with task-tree planning for Windows tasks. Hierarchical prompting~\cite{lo2023hierarchical} and reasoning-in-the-loop agents like ReAct~\cite{yao2023react}, MM-ReAct~\cite{yang2023mm}, and Reflexion~\cite{shinn2024reflexion} improve robustness through stepwise feedback. UFO~\cite{zhang2024ufo} introduces modular planning with app-specific agents, and VisionTasker~\cite{song2024visiontasker} mitigates hallucinated steps using user demonstrations. Recent surveys~\cite{nguyen2024gui,sager2025ai} identify global planning followed by local grounding as an emerging standard.

Recent VLM-based systems, including UI-TARS~\cite{qin2025ui,hong2024cogagent,li2024ferret} and commercial tools~\cite{amazon2025novaact,anthropic2024computeruse,deepmind2024projectmariner}, expand model size and platform coverage. MoTIF~\cite{burns2022dataset} supports comparative evaluation across these approaches.

Nonetheless, three practical challenges remain. First, multi-screenshot prompts often exceed the model’s context window, leading to out-of-memory failures~\cite{gao2023assistgui,li2024ferret}. Second, per-step vision encoding incurs growing latency, particularly in tasks exceeding 20 subtasks~\cite{amazon2025novaact,deepmind2024projectmariner}. Third, plan reliability significantly degrades beyond 15 actions, revealing limitations in maintaining long-horizon coherence~\cite{shinn2024reflexion,yao2023react,zhang2024ufo}.

\subsection{GUI Log-Mining and Robotic-Process Automation}
Complementing the interaction-log theme introduced in §1, traditional RPA systems approached GUI control through log mining and macro scripting. Enterprise recorders captured user interactions—clicks, key-presses, window transitions—to reconstruct process graphs later converted into parameterized routines~\cite{choi2022enabling}. Large-scale datasets such as RICO~\cite{deka2017rico} and the web-scale ETNA crawler~\cite{riva2021etna} extended coverage, while tools like ParamMacros exposed argument slots to support partial generalization~\cite{krosnick2022parammacros}. Some mobile systems adapted these techniques to voice-driven shortcuts~\cite{arsan2021app}.

To reduce manual rule engineering, recent work has begun to integrate large language models. Automatic Macro Mining demonstrates that LLMs can cluster millions of interaction traces and assign natural-language labels to each cluster, cutting curator workload by over 60\% ~\cite{huang2024automatic}. Generative models also improve variant discovery in process graphs, though struggle on rare transitions\cite{fani2023llms}. On mobile, fine-tuned decoders translate free-form commands into gesture sequences with 82\% exact match accuracy~\cite{li2020mapping}, and interactive wrappers refine execution via dialogue~\cite{jiang2025iluvui}.TaskLAMA~\cite{yuan2024tasklama} stresses these models with long, interleaved instructions to reveal edge cases beyond the reach of classical miners.

However, these fundamental limits persist: script stability degrades under UI layout changes, as shown in ParamMacros~\cite{krosnick2022parammacros} and RICO-based replay failures~\cite{deka2017rico}; and compiled plans remain static at runtime, lacking the self-repair capabilities found in reasoning-based agents~\cite{huang2024automatic,fani2023llms,arsan2021app}.

\subsection{Integrating Logs and Demonstrations into LLM-Based Agents}
Programming-by-demonstration (PBD) systems such as SUGILITE, APPINITE, and PUMICE~\cite{li2017sugilite,li2018appinite,li2019pumice} enabled end users to automate tasks by combining GUI interactions with natural language. Systems like Kite~\cite{li2018kite} and VASTA~\cite{sereshkeh2020vasta} extended this approach to mobile environments, offering multi-modal grounding through demonstration.

While PBD approaches are more adaptive than rule-based scripting, they (i) struggle with ambiguity resolution and task generalization across dynamic layouts~\cite{huang2024automatic,yuan2024tasklama}, and (ii) rely heavily on template-based mappings, limiting plan reuse or transfer across apps and users.
To address these issues, recent work has attempted to combine task logs or UI understanding with LLMs. Task mining methods such as Redis et al.\cite{redis2024skill} inject structural priors (e.g., process graphs), while systems like Zhang et al.\cite{zhang2024dynamic} achieve strong action prediction accuracy by modeling GUI states.

However, these LLM-augmented approaches still face critical limits:
(i) most rely on static plan construction and cannot adapt at runtime when the model hallucinates or misses context~\cite{huang2024automatic,yuan2024tasklama,redis2024skill};
(ii) accurate single-step grounding, as in~\cite{zhang2024dynamic}, often ails to chain into robust long-horizon plans due to the absence of explicit planning modules.

Log2Plan addresses these gaps by unifying symbolic memory and retrieval-augmented generation. It (a) derives GUI interaction-to-action mappings from past demonstrations, (b) performs real-time semantic retrieval over these mappings using user commands, and (c) constructs a high-level plan before grounding and execution. Unlike prior systems that replay logs or apply direct action mappings, Log2Plan dynamically composes plans conditioned on both past behavior and present intent - without fine-tuning. This enables robust generalization across diverse, unseen GUI tasks.
We now present the planning and execution pipeline in Section \ref{sec:3}.

\section{Task Mining and Log-Guided Semantic Retrieval}
\label{sec:3}
Log2Plan turns raw desktop traces into reusable knowledge through a tightly coupled approach consisting of structured task mining followed by semantic retrieval and reuse (described in Sections 3.1 and \ref{sec:3.2}, respectively). The same pipeline operates in two modes: offline processing, where it structures 21 hours of interaction data (120 logs collected over three weeks) into searchable task patterns, and online retrieval, where it combines patterns from multiple logged sessions to enable GlobalPlanner (Section \ref{sec:4}) to execute complex, multi-application workflows that no single logged interaction contains. Unlike flat record-and-replay scripts, our hierarchy of Task Blocks, Individual Tasks, and Task Groups abstracts away superficial UI changes and scales to long-horizon goals. 
\subsection{Structured Task Mining from User Logs}
\label{sec:3.1}
\begin{figure*}[ht!]
    \centering
    \includegraphics[width=\linewidth]{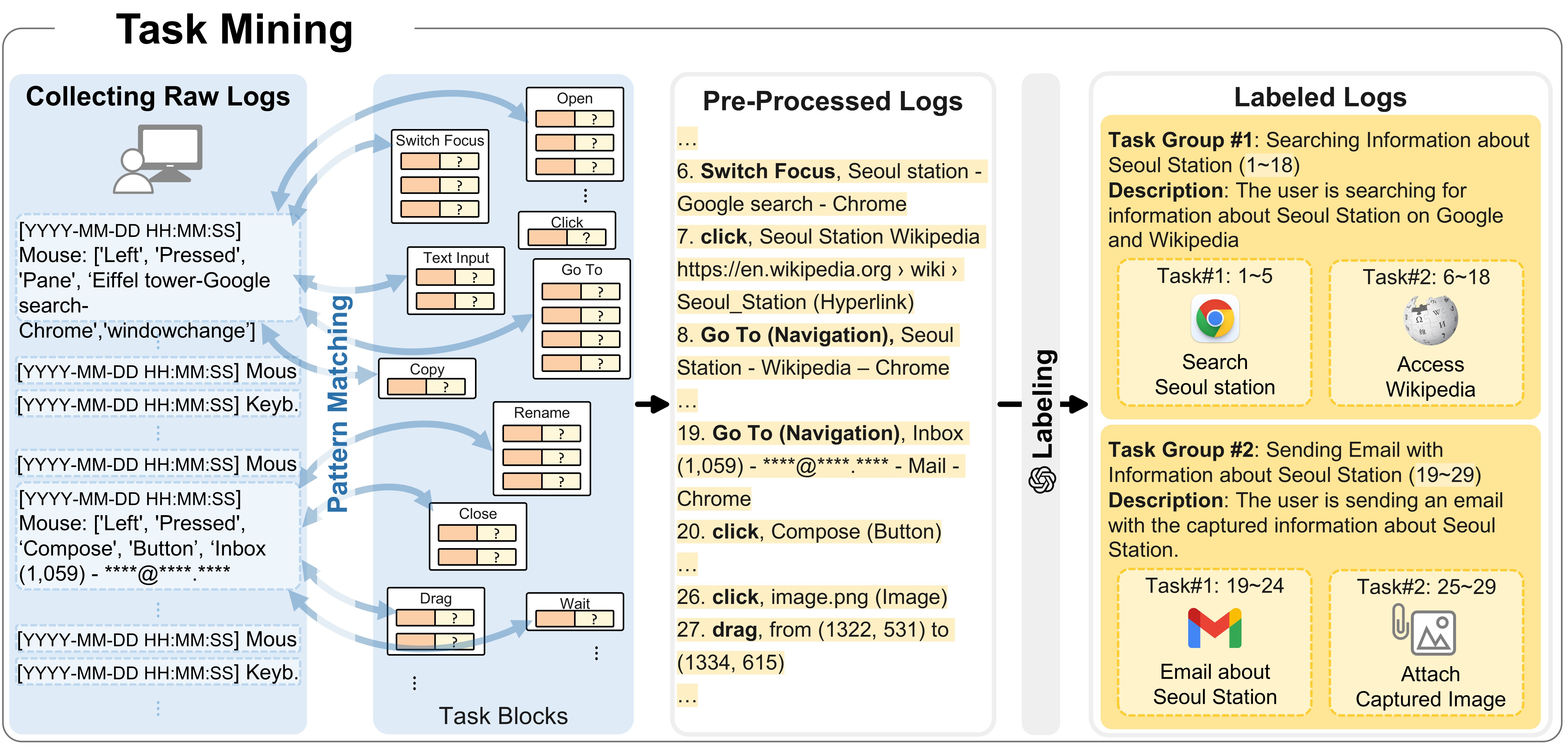}
    \caption{Overview of the task mining process. The process includes collecting raw logs, applying pattern matching to transform raw logs into structured task blocks, and labeling them as meaningful task groups. Each task group follows the ENV[environment] ACT[action] Title format, with accompanying descriptions and numbered subtasks.}
    \label{fig:task_mining}
\end{figure*}

Task mining bridges low-level traces and high-level automation by converting raw desktop events into modular, reusable knowledge units. The proposed task mining process follows two ordered steps:
\begin{enumerate}
    \item \textit{Pre-processing}: Transforming raw logs into structured task blocks by mapping sequences of low-level events to 19 predefined high-level GUI events
    \item \textit{Hierarchical Labeling}: Organizing processed logs into Task groups with ENV, ACT, Description structure
\end{enumerate}

\paragraph{Pre-processing} Raw desktop events including mouse actions, keyboard inputs, and window context events are continuously intercepted at the operating-system layer and recorded with timestamps and associated object information to form sequential user activity logs. These raw logs are brittle because identical intents surface as different UI, captions, or timing patterns. We therefore map variable-length event sequences to 19 predefined high-level GUI events such as ``Switch Focus'', ``Text Input'', and ``Go To'' using pattern-matching rules detailed in Appendix~\ref{appendix-A}.The result is a list of Task Blocks, each formatted into a standardized structure, where\texttt{user\_assist} indicates whether manual execution by the user is required:
\begin{equation*}
[\texttt{high-level event, objects}]
\end{equation*}
An example of such pre-processed logs is shown in Figure~\ref{fig:task_mining}.

\paragraph{Hierarchical Labeling} To transcend task-specific scripts and support novel task combination, we organize the Task Blocks into a four-level hierarchy using GPT-4o\cite{openai2024hello} model. The hierarchy consists of ENV[environment/platform] indicating the executing environment or platform (e.g., web/chrome-google map, local/FileExplorer), ACT[action category/specific action] representing the action category and specific action (e.g., search/text input, navigation/open browser), a Title summarizing the overall goal, and description providing a one sentence natural language summary, as illustrated in Figure~\ref{fig:task_mining}. GPT-4o model segments the block stream into numbered individual Tasks and rolls related ones into a Task Group that pursues a coherent goal.

By mining repetitive patterns and encoding them with this ENV/ACT/Title/Description schema, our system creates structured embeddings that enable more precise categorization and semantic understanding of logged interactions. This hierarchical organization allows the system to accurately capture user intent from historical logs, significantly improving automation accuracy and enabling intelligent pattern reuse. The detailed operational principles of frequency-based extraction, similarity-based adaptation, and context-aware reasoning are demonstrated through a comprehensive case study in Section~\ref{sec:CaseStudy}

\subsection{Semantic Retrieval and Reuse of User Task Groups}
\label{sec:3.2}
We embed each Task Group’s concatenated ENV/ACT/Title labels and Description using OpenAI’s text-embedding-ada-002 model\cite{openai2022embedding} and store the resulting vectors for similarity search. When a user issues a command, the query is first broken down into ENV/ACT/Title format to enhance similarity assessment accuracy, then embedded in the same space and compared using cosine similarity with a staged diversity selection strategy to retrieve k=9 Task groups, following the effective retrieval-augmented generation approach of \cite{huang2024cosent}. This strategy first selects the top 3 most similar groups, then iteratively excludes the most similar candidates to each selected unit and chooses 3 more, repeating this process to ensure diverse coverage while maintaining relevance. For every individual task in the generated plan, we then fetch the top-2 most similar individual tasks from the retrieved groups, enabling precise task-level matching. These retrieved patterns are incorporated into Globalplanner’s (see Section~\ref{sec:Global}) prompt to generate a new set of high level tasks. The dynamic retrieval approach allows the system to adapt to novel task combinations by selecting and combining relevant patterns from different task groups, significantly reducing the search space while maintaining behavioral relevance. This multi-source pattern combination enables the system to handle complex, multi-step user queries even when no single logged session contains the exact sequence, executing sophisticated automation workflows that exceed the scope of any individual logged interaction.

\section{Planning}\label{sec:4}
To bridge the gap between high-level user intent and low-level executable actions, our system adopts a two-level planning structure consisting of a GlobalPlanner and a LocalPlanner (described in Sections \ref{sec:Global} and \ref{sec:Local}, respectively). The GlobalPlanner translates a user’s natural language command into a sequence of high-level events(actions) by referring to semantically similar task groups extracted from recorded user activity logs. This approach enables user-adaptive planning and enhances the system’s ability to interpret GUI environments with limited prior knowledge, even without direct access to current GUI information. Following this, the LocalPlanner transforms each high-level action into a task block in the Appendix~\ref{appendix-A}, which is a sequence of low-level GUI actions tailored to the current screen state. This hierarchical structure---from natural language command to high-level actions, and finally to low-level actions---allows the system to maintain both generalization and precision across various desktop automation scenarios.

\subsection{Global Planner}\label{sec:Global}
\begin{figure}[t]
    \centering
    \includegraphics[width=\linewidth]{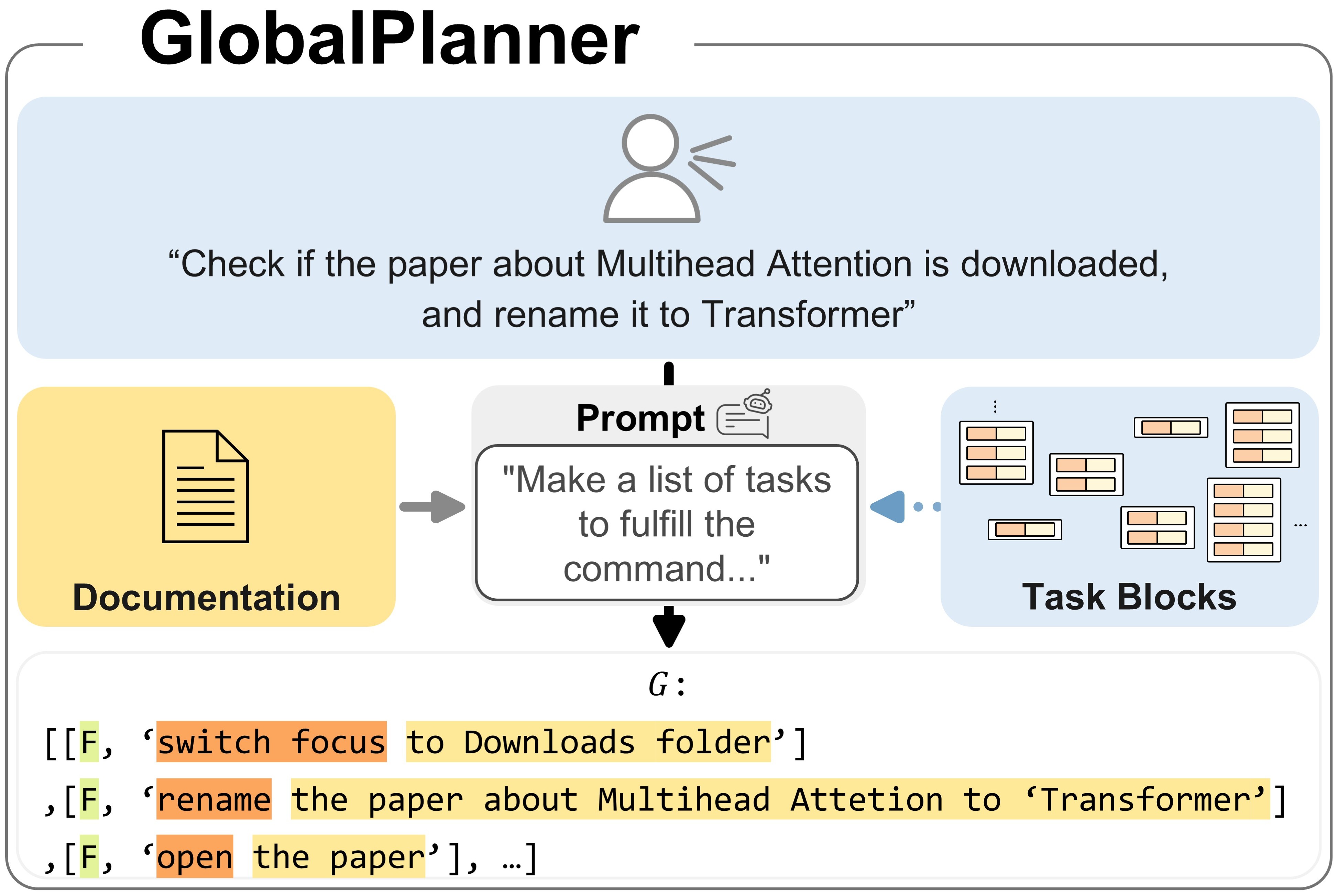}
    \caption{The GlobalPlanner generates a high-level task list from a natural language command using prompt-based reasoning with task reference retrieval.}
    \label{fig:global_planner}
\end{figure}
The GlobalPlanner decomposes a user command expressed in natural language into a task list, denoted as $G = [g_1, g_2, \dots , g_N]$, where $g_i$ is an individual task corresponding to a high-level action and $N$ is the total number of tasks. The structure of each task $g_i$ is defined as:
\begin{align*}
    g_i=\texttt{[user-assist, high-level action, object]}
\end{align*}
which is consistent with the standardized log format described in the task mining process. The high-level action is selected from one of the 17 predefined log event types, ensuring that tasks are defined in a consistent and automation-ready manner.

Using GPT-4o\cite{openai2024hello} as the baseline model, the GlobalPlanner outlines the high-level steps required to perform the user command and defines a global plan that specifies the execution sequence of all tasks. Each task $g_i$ consists of a user-assist flag (T / F), an action (event), and corresponding objects. These objects denote the GUI components (e.g., buttons, file entries, input fields) that are directly involved in or affected by the action, enabling fine-grained control over task execution. This structure enables systematic and interpretable decomposition of complex commands.

In generating the task list, the GlobalPlanner references semantically similar task groups previously extracted and labeled during the task mining process. These task groups are retrieved through vector similarity search based on the current user query. By incorporating knowledge from similar past tasks, the GlobalPlanner can better understand the intent of the user and adapt the global plan to unfamiliar or dynamic GUI environments. This not only increases the robustness of task list generation but also improves generalization, as shown in Figure~\ref{fig:Example},  enabling zero-shot planning capabilities even in previously unseen application contexts.

The GlobalPlanner thus serves as a bridge between high-level user intent and low-level executable actions by generating a structured, semantically grounded task list that guides the rest of the automation pipeline. If the resulting task list is later found to be incomplete or infeasible during execution, the LocalPlanner, introduced later, can refine or revise the plan accordingly.

\subsection{Local Planner}\label{sec:Local}
\begin{figure*}
    \centering
    \includegraphics[width=\textwidth]{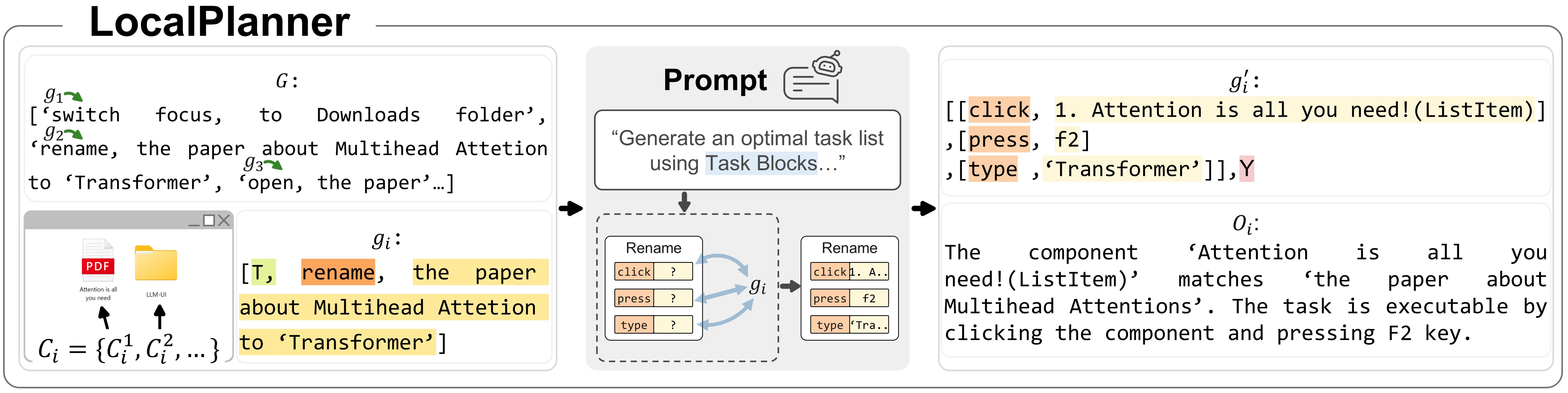}
    \caption{Key inputs, outputs, and structure of the LocalPlanner module: The LocalPlanner interprets the current screen state based on the current task and component list. It then provides an assessment of task feasibility, observation, and a corresponding task list for the current situation.}
    \label{fig:Local-planner}
\end{figure*}

The LocalPlanner revises the current task $g_i$ by considering the state of the GUI environment prior to its execution. As illustrated in Figure~\ref{fig:Local-planner}, it receives as input the task list $G$, the current task $g_i$, and the component dictionary $C_i$. For a given task $g_i$, the dictionary $C_i$ consists of a set of related components, denoted as $C_i=\{C^1_i, C^2_i, \dots, C^j_i,\dots\}$. Each component $C_i^j$ is represented as follows:
\begin{align*}
C_i^j = \texttt{[name, control-type, position)]}
\end{align*}

The component dictionary is constructed using \texttt{PyWinAuto}, a GUI automation library that enables inspection of GUI controls and programmatic interaction with Windows applications. This allows the system to access detailed information about GUI components and perform low-level actions on them as needed for task execution.

Based on these inputs, the LocalPlanner identifies the high-level actions associated with the task and selects the most contextually appropriate task block from a predefined set corresponding to that high-level actions. The selected task block is then adapted to the user’s intent by modifying it according to the current GUI state and the specific context of the task.\\

For the task `Rename the paper about Multihead Attention to Transformer', the selected `Rename' task block is as below:
\begin{align*}
&\texttt{Rename Task Block: \{click + object, press + f2,}\\
&\texttt{type + object, press enter\}}\\
\end{align*}
To align the generic task block with the specific user intent, each \texttt{object} is replaced with a contextually relevant entity from the GUI environment. For instance, if the task can proceed, task block is modified into the following low-level action sequence:\begin{align*}
&\texttt{\{click `1. Attention is all you need! (ListItem)',}\\
&\texttt{press `f2', type `Transformer', press enter\}}\\
\end{align*}

The execution flag $E_i$ is set to \texttt{Y}, indicating that the task is executable. In this case, the updated task block is passed to the Execution module. Simultaneously, the LocalPlanner generates the observation $O_i$, which provides an explanation of the update and the reasoning process behind the modification of the task block $g'_i$.

\subsection{Execution}
The task block $g'_i$, consisting of a sequence of low-level actions generated and refined by the LocalPlanner, is executed by the Execution module. This module translates each action in the task block into executable instructions to perform the corresponding low-level actions within the GUI environment. The types of GUI controls supported by the Execution module include \texttt{press}, \texttt{type}, \texttt{click}, \texttt{double click}, \texttt{right click}, \texttt{drag}, \texttt{scroll}, and \texttt{focus}.

To carry out these operations, the Execution module leverages both \texttt{PyWinAuto} and \texttt{PyAutoGUI}, enabling reliable automation of a wide range of control behaviors across Windows applications. If a low-level action requires user intervention--such as selecting a specific item or entering sensitive information--a pop-up window is displayed to prompt the user for input. Whether such intervention is necessary is determined by examining the user-assist bit embedded in $g_i$. By leveraging manual execution when necessary, the system can resolve ambiguities in user intent while maintaining a high level of automation, thereby enhancing overall task execution efficiency.
\vspace{-20pt}
\section{Evaluation and Discussion}
\subsection{Evaluation Setup}
\begin{figure*}
    \centering
    \includegraphics[width=0.8\linewidth]{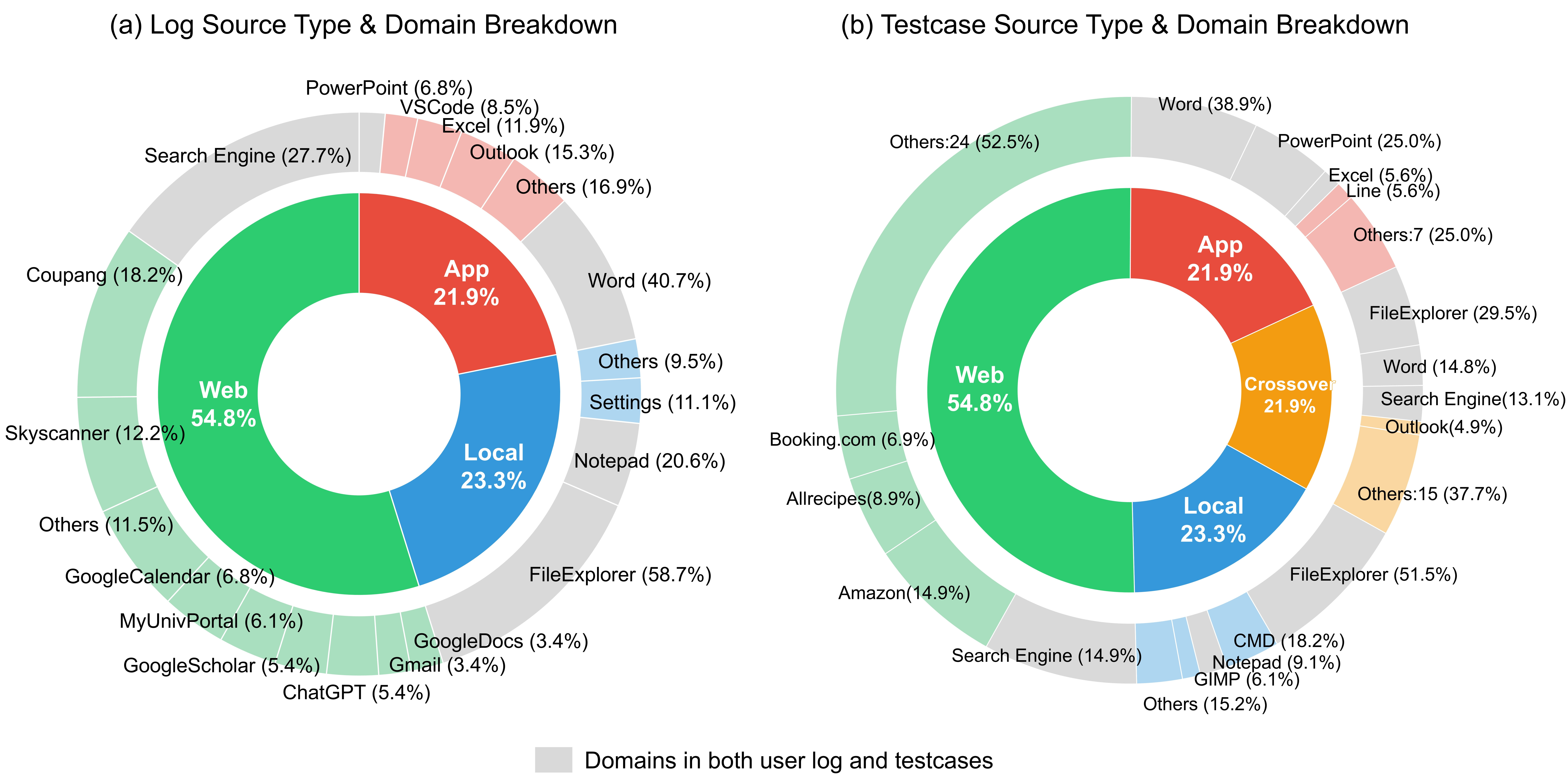}
    \caption{Domain distributions of user logs (left) and testcases (right). Inner rings indicate the proportion of source types: Web, Local, App, and Crossover; outer rings show domain-level breakdowns. Domains shared across both sets are shown in gray.}
    \label{fig:domaincomp}
\end{figure*}

We evaluate the effectiveness of Log2Plan using a total of 200 real-world GUI automation tasks. The evaluation set consists of 100 tasks collected in-house, randomly sampled 50 tasks from the SkyVern~\cite{skyvern2025introduction} dataset(639 total) ScreenAgent~\cite{niu2024screenagent} dataset(70 sessions total) each. These tasks cover a wide range of environments, including local desktop applications(Local), web-based platforms(Web), productivity software (App), and cross-application workflows (Crossover). Each task was annotated with its primary execution type based on the dominant context. The distribution of testcases is shown in of Figure~\ref{fig:domaincomp}(b).

To support task mining, we collected approximately 20 hours of GUI activity and segmented the data based on user behavior and environment. Using a one-hour inactivity threshold, we first divided the timeline into 12 distinct user sessions. Each session was then further split by execution environment—such as domain or application type—yielding a total of 141 interaction log files. The distribution of these environments is visualized in Figure~\ref{fig:domaincomp} (a), grouped into Local, Web, and App categories.

Notably, the domains of logs and testcases are intentionally imbalanced: only a subset of environments (shown in gray) appear in both. For instance, tools like Chrome or File Explorer are shared, while many Web testcases—spanning 28 domains, with 24 classified under `Others', do not exist in the collected logs. This setting enables us to evaluate how well task mining generalizes to unseen domains.

We compare Log2Plan with four baselines: (1) Log2Plan w/o TM, which removes task retrieval; (2) ReAct, an LLM-based step-by-step planner with GUI dictionary access and coupled with the same execution backend as ours; (3) UFO2~\cite{zhang2025ufo2}, a Windows automation agent; and (4) UI-TARS~\cite{qin2025ui}, a visual planner–executor using screen-based grounding. All models share a unified GUI action interface of atomic operations. We report total/partial success rates and task completion time. Further details appear in Appendix~\ref{appendix-B}.

\subsection{Evaluation of End-to-End Task Execution}
\label{evaluation-results}
\begin{table}
    \centering
    \begin{tabular}{lccc}
        \toprule
                                &\multicolumn{2}{c}{Metrics} \\\cmidrule{2-3}
        \multirow{2}{*}{Method} &   Execution Time $\downarrow$   & Success Rate $\uparrow$\\
                                &   (sec/task)& (\%) \\
        \midrule
        React-style Planner    &   28.6&  18.0\\
        UFO2~\cite{zhang2025ufo2}  &  118.2&   46.5\\
        \midrule
        Log2Plan w/o TM &  44.2&  28.0\\
        Log2Plan        &  80.0 &   61.7\\
        \bottomrule
    \end{tabular}
    \caption{Comparison of average execution time per task (sec) and success rate (\%) across four models: ReAct-style Planer, UFO, Log2Plan without TM(Task Mining), and ours, Log2Plan.}
    \label{tab:success_rate_and_time}
\end{table}

\begin{figure}
    \centering
    \includegraphics[width=0.9\linewidth]{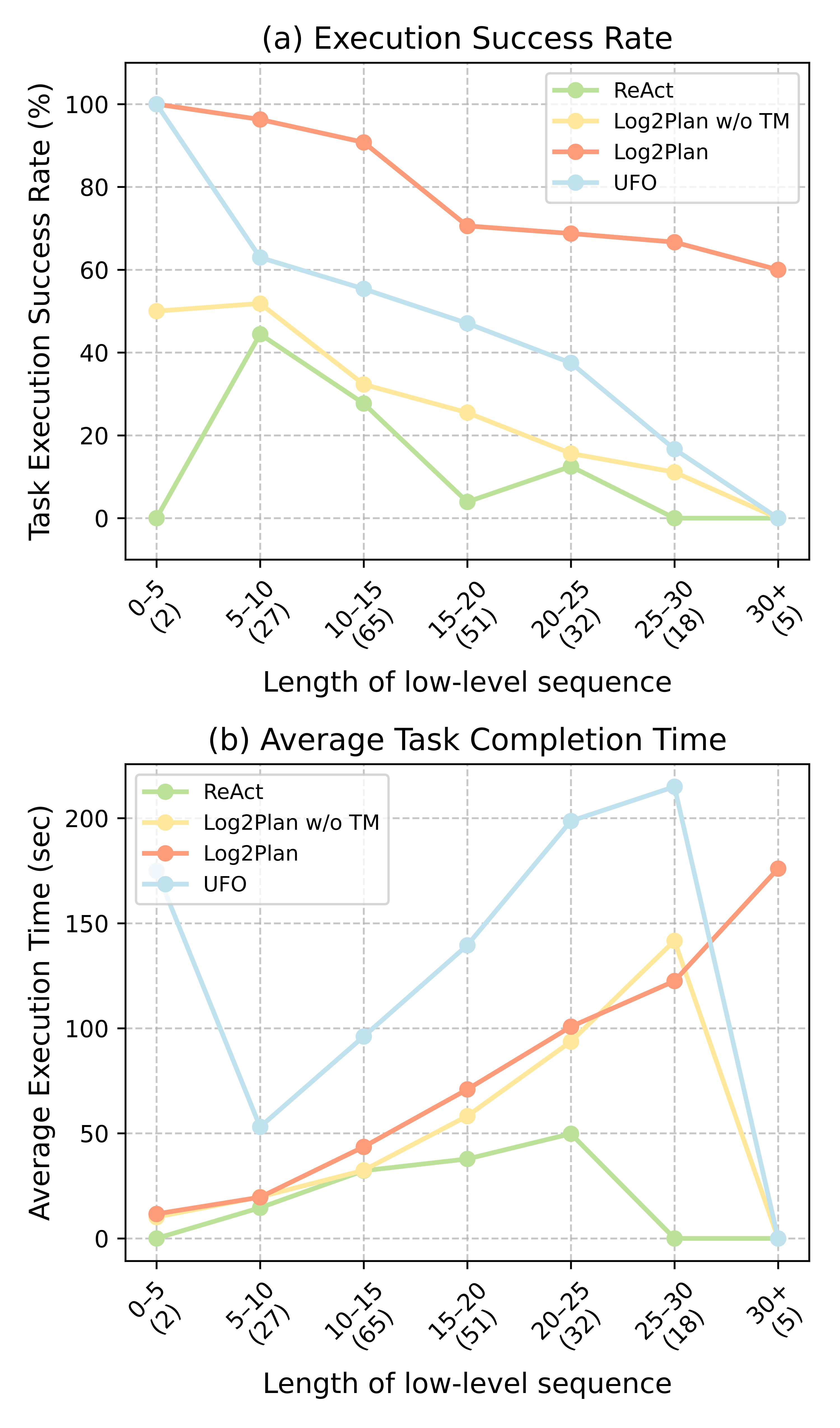}
    \caption{Comparison of four models ( UFO, ReAct, Log2Plan without Log data, and Log2Plan) across two metrics: (a) execution success rate by low-level action count, and (b) execution time by low-level action count. The graphs illustrate how model performance varies with task complexity and efficiency.}
    \label{fig:qualitative_comp}
\end{figure}

We assess four agents—Log2Plan, Log2Plan w/o TM, UFO, and ReAct—on 200 GUI automation tasks, grouped by low-level action sequence length to analyze scalability. Results are shown in Figure~\ref{fig:qualitative_comp} and Table~\ref{tab:success_rate_and_time}.

Log2Plan achieves the highest success rate (80.0\%) while maintaining reasonable completion time (61.7 sec), demonstrating both robustness to domain shift and efficiency. The ablated version drops to 28.0\%, confirming the critical role of task retrieval in supporting full execution. By leveraging past demonstrations, Log2Plan enhances global planning performance significantly. UFO and ReAct yield lower success rates at 46.5\% and 18.0\%, respectively.

Performance gaps widen with task complexity. In the 25–30 and 30+ action bins, Log2Plan maintains over 60\% success, whereas other models, including the ablated variant—fail to generalize to longer sequences. ReAct’s step-by-step visual grounding without structural guidance limits its effectiveness, while UFO suffers from decreased stability as task length increases, suggesting coordination challenges for longer workflows. Execution time comparisons support these observations: Log2Plan remains more efficient than UFO and avoids timeout failures frequently encountered by ReAct.

Overall, these results highlight the architectural strengths of Log2Plan in enabling scalable, reliable end-to-end execution across complex and unfamiliar environments.

\subsection{Evaluation of Average Subtask Completion Rate}
\begin{table}[t]
    \centering
    \begin{tabular}{lc}
        \toprule
        Method & Avg. Subtask Completion Rate (\%) $\uparrow$ \\
        \midrule
        UI-TARS~\cite{qin2025ui}       & 36.9 \\
        UFO2~\cite{zhang2025ufo2}       & 62.8 \\
        Log2Plan w/o TM                & 58.7 \\
        Log2Plan (ours)               & \textbf{93.4} \\
        \bottomrule
    \end{tabular}
    \caption{Average subtask completion rate across 200 test cases. Each value denotes the mean percentage of subtasks successfully completed within a full task sequence.}
    \label{tab:subtask_completion_rate}
\end{table}
To further assess model robustness beyond full execution success, we evaluate how much of each task sequence was completed before failure. Table~\ref{tab:subtask_completion_rate} reports the average subtask completion rate—i.e., the proportion of subtasks successfully carried out within each task.

Log2Plan achieves the highest completion rate at 93.4\% across 200 testcases, indicating consistent progress even in partially failed executions. Notably, Log2Plan w/o TM performs competitively at 58.7\%, suggesting that its hierarchical architecture alone contributes significantly to execution stability.  In contrast, UI-TARS, a vision-language-based step-by-step planner, shows a markedly lower rate of 36.9\%, often failing early during page load or initial UI parsing. This underscores the limitations of vision-only grounding in handling long or unfamiliar workflows. UFO, despite adopting a two-level structure, achieves 62.8\%, likely due to its under-specified planning transitions, which can lead to inconsistent subgoal completion.

While the overall success rate of Log2Plan without task retrieval remains lower, its partial execution performance is comparatively strong. In several challenging web and app scenarios, where UI-TARS and UFO often failed to launch or interact with the initial UI element, both Log2Plan variants were able to complete multiple high-level actions before termination. This indicates that even without task mining, Log2Plan's structured planner supports more sustained progress than other baselines. These findings collectively highlight the robustness of Log2Plan’s two-level planning, which enables progressive and resilient task execution under distribution shifts or partial failures.

\subsection{Case Study: Generalization to GUI Environments with Limited Prior Knowledge}
\label{sec:CaseStudy}
\begin{figure*}
    \centering
    \includegraphics[width=\linewidth]{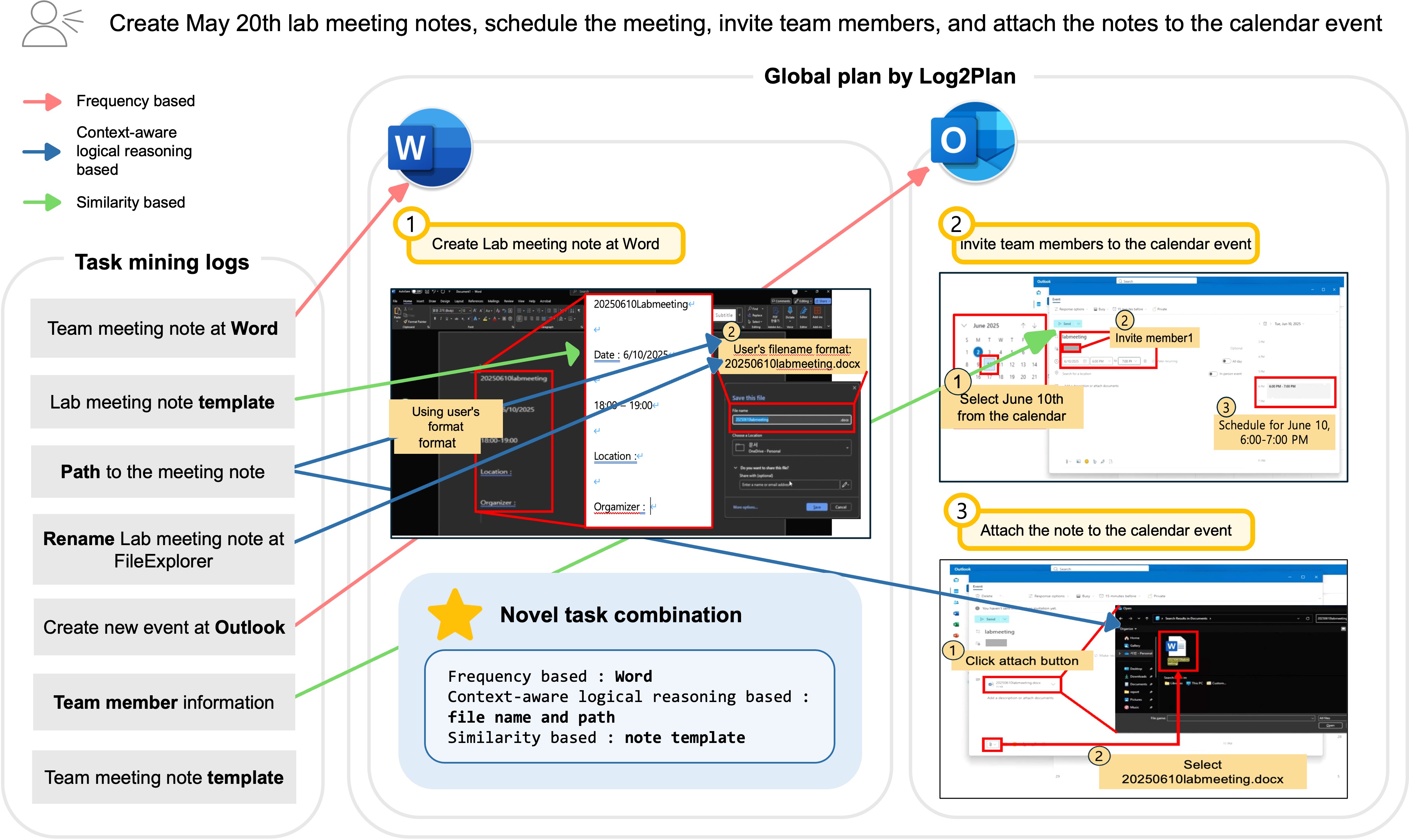}
    \caption{Log2Plan's Three-Principle Approach to Complex Task Automation. The system retrieves and combines patterns from multiple logged sessions using frequency-based extraction (red), similarity-based adaptation (green), and context-aware logical reasoning (blue) to execute a sophisticated multi-application workflow. The central combination box shows how these principles enable novel task synthesis that exceeds any individual logged interaction.}
    \label{fig:Example}
\end{figure*}

Figure~\ref{fig:Example} shows how Log2Plan leverages its three-principle approach to handle a complex, multi-step automation task. The task is to create lab meeting note using template at Word, schedule a meeting, invite team members, and attaching note to a calendar event at Outlook—a sophisticated workflow requiring coordination across multiple applications.

Log2Plan employs three distinct principles for retrieval. For the first action "Create Lab meeting note at Word", Log2Plan employs frequency-based extraction to select Word as the primary application from "Team meeting note at Word", leveraging the most commonly used environment for document creation. Similarity-based adaptation retrieves the "Lab meeting note template" to provide the appropriate document structure, while context-aware logical reasoning utilizes "Path to the meeting note" and "Rename Lab meeting note at FileExplorer" to determine the proper file naming and storage location.

For the subsequent actions, the mined logs enable the system to easily adapt to specific application buttons and UI elements that would otherwise be impossible to execute without the mined logs. The logged interactions provide crucial guidance for navigating Outlook's calendar interface, locating invitation buttons, and understanding file attachment mechanisms—actions that would fail without the mined logs.

This multi-source approach enables the system to construct a novel task combination, where no single logged session contains this exact sequence. The case demonstrates how Log2Plan transcends simple replay mechanisms by dynamically combining behavioral patterns from different contexts through its three complementary principles, achieving cross-application coordination that maintains task coherence throughout the automation workflow.

\section{Conclusion}
This paper introduced Log2Plan, a two-level GUI automation framework that addresses key limitations of existing Programming by Demonstration systems, including low adaptability, fragile planning, and poor handling of dynamic GUI environments. Unlike prior approaches reliant on brittle sequences or rigid templates, Log2Plan separates high-level planning from low-level execution and uses structured task mining from user logs to build reusable and personalized automation flows. It integrates a GlobalPlanner that interprets user commands via log-based retrieval and a LocalPlanner that grounds tasks into executable actions based on the current GUI context. This hierarchical structure enhances robustness to UI changes and generalization to unseen interfaces. We evaluated Log2Plan on 200 real-world desktop automation tasks. The results show higher success rates and faster execution times than the baseline models. In particular, Log2Plan maintains stable performance even for tasks with more than 20 low-level actions, demonstrating strong scalability. These findings suggest that combining structured behavior mining with modular planning enables more reliable, adaptive, and personalized GUI automation. Using user behavior and updating task knowledge over time, Log2Plan adapts to individual usage patterns, offering a promising path toward general-purpose automation agents that effectively bridge user intent and execution.

However, several limitations remain. As user logs accumulate over time, the system tends to retain redundant or low-significance task files, which can dilute retrieval quality and introduce unnecessary overhead during planning. This issue becomes more pronounced in long-term deployments, where similar or noisy task entries are frequently repeated. Additionally, Log2Plan currently identifies GUI components based solely on their textual labels, making it difficult to distinguish between visually similar elements when their names are ambiguous or lack semantic richness. For example, differentiating the content or context of an image file named dog.png. To address these challenges, future work will explore log filtering techniques to maintain a high-quality task repository and incorporate lightweight visual recognition within component viewports to improve semantic disambiguation beyond textual identifiers.

\begin{acks}
This work was supported by Institute of Information \& communications Technology Planning \& Evaluation (IITP) grant funded by the Korea government(MSIT) (No.2022-0-00025, Development of soft-suit technology to support human motor ability) and the National Research Foundation of Korea(NRF) grant funded by the Korea government(MSIT) (No. 2022H1D8A3037394).
\end{acks}

\bibliographystyle{ACM-Reference-Format}
\bibliography{references}

\appendix

\section{Task Dictionary}\label{appendix-A}
The Task Dictionary below summarizes how each high-level GUI action is translated into a sequence of low-level events (Table~\ref{appdx:task_dictionary}). These mappings are used during both task mining and local planning. Each entry represents a reusable automation pattern.

\begin{table*}
\centering
\begin{tabular}{clllc}
\toprule
\multirow{2}{*}{High-level Action}  &\multicolumn{4}{c}{Low-Level Event Sequence} \\\cmidrule{2-5}
&\multicolumn{1}{c}{Step 1}&\multicolumn{1}{c}{Step 2}&\multicolumn{1}{c}{Step 3}&\dots\\
\midrule
\multicolumn{1}{l}{Text Input}          &\texttt{click, text input field}   &\texttt{type, text input field}            &\texttt{press, enter}&--\\
\multicolumn{1}{l}{Text Input}          &\texttt{click, text input field}   &\texttt{type, text input field}            &\texttt{press, tab}&--\\
\multicolumn{1}{l}{Click}               &\texttt{click, target object}      &--&--&--\\
\multicolumn{1}{l}{Doubleclick}         &\texttt{doubleclick, target object}&--&--&--\\
\multicolumn{1}{l}{Rightclick}          &\texttt{rightclick, target object} &\texttt{move focus to child window(menu)}  &--&--\\
\multicolumn{1}{l}{Drag}                &\texttt{mousedown, target object}  &\texttt{mouseup, destination}              &--&--\\
\multicolumn{1}{l}{Scroll}              &\texttt{focus on, target window}   &\texttt{scroll, 2/3$\cdot$windowHeight}    &--&--\\
\multicolumn{1}{l}{Scroll}              &\texttt{mousedown, scroll bar}     &\texttt{mouseup, 2/3$\cdot$windowHeight}   &--&--\\
\multicolumn{1}{l}{Press}               &\texttt{press, keys}               &--&--&--\\
\multicolumn{1}{l}{Open}                &\texttt{press, win}                &\texttt{type, target}                      &\texttt{press, enter}&--\\
\multicolumn{1}{l}{Open}                &\texttt{doubleclick, target}       &--&--&--\\
\multicolumn{1}{l}{Open}                &\texttt{click, search bar}         &\texttt{type, target}                      &\texttt{press, enter}&--\\
\multicolumn{1}{l}{Close}               &\texttt{focus on, title bar}       &\texttt{doubleclick, close button}         &--&--\\
\multicolumn{1}{l}{Close}               &\texttt{press, alt+f4}             &--&--&--\\
\multicolumn{1}{l}{Switch Focus}        &\texttt{press, alt+esc}            &--&--&--\\
\multicolumn{1}{l}{Switch Focus}        &\texttt{move focus to task bar}    &\texttt{press, win+tab}                    &\texttt{press, arrow keys}&\dots\\
\multicolumn{1}{l}{Switch Focus}        &\texttt{press, ctrl+t}             &\texttt{type, url}                         &\texttt{press, enter}&--\\
\multicolumn{1}{l}{Switch Focus}        &\texttt{press, ctrl+shift+tab}     &--&--&--\\
\multicolumn{1}{l}{Go To (Navigation)}  &\texttt{press ctrl+l}              &\texttt{type, url}                         &\texttt{press, enter}&--\\
\multicolumn{1}{l}{Go To (Navigation)}  &\texttt{click, hyperlink}          &--&--&--\\
\multicolumn{1}{l}{Go To (Navigation)}  &\texttt{click, dropdown button}    &\texttt{press, arrow keys}                 &\texttt{press, enter}&--\\
\multicolumn{1}{l}{Go To (Navigation)}  &\texttt{press, alt}                &\texttt{press, arrow keys}                 &\texttt{press, enter}&--\\
\multicolumn{1}{l}{Save}                &\texttt{press, ctrl+s}             &--&--&--\\
\multicolumn{1}{l}{Copy}                &\texttt{click, target}             &\texttt{press, ctrl+c}                     &--&--\\
\multicolumn{1}{l}{Paste}               &\texttt{press, ctrl+v}             &--&--&--\\
\multicolumn{1}{l}{Delete}              &\texttt{click, target}             &\texttt{press, ctrl+d}                     &--&--\\
\multicolumn{1}{l}{Rename}              &\texttt{click, target object}      &\texttt{press, f2}                         &\texttt{type, name}&\dots\\
\multicolumn{1}{l}{Login}               &\texttt{click, login button}       &\texttt{click, id field}                   &\texttt{type, user id}&\dots\\
\multicolumn{1}{l}{Login}               &\texttt{click, login button}       &\texttt{click, id field}                   &\texttt{type, user id}&\dots\\
\multicolumn{1}{l}{Repeat}              &\texttt{repeat, task\#, on object} &--&--&--\\
\multicolumn{1}{l}{Wait}                &\texttt{wait, screen change}       &--&--&--\\
\bottomrule
\end{tabular}
\caption{Mapping each high-level event to a corresponding sequence of low-level events. This serves as foundational knowledge for task mining, where combinations of high-level events are used to diverse user behaviors across GUI operations.}
\label{appdx:task_dictionary}
\end{table*}

\section{Experiment Datails}\label{appendix-B}
\begin{figure*}
    \centering
    \includegraphics[width=\linewidth]{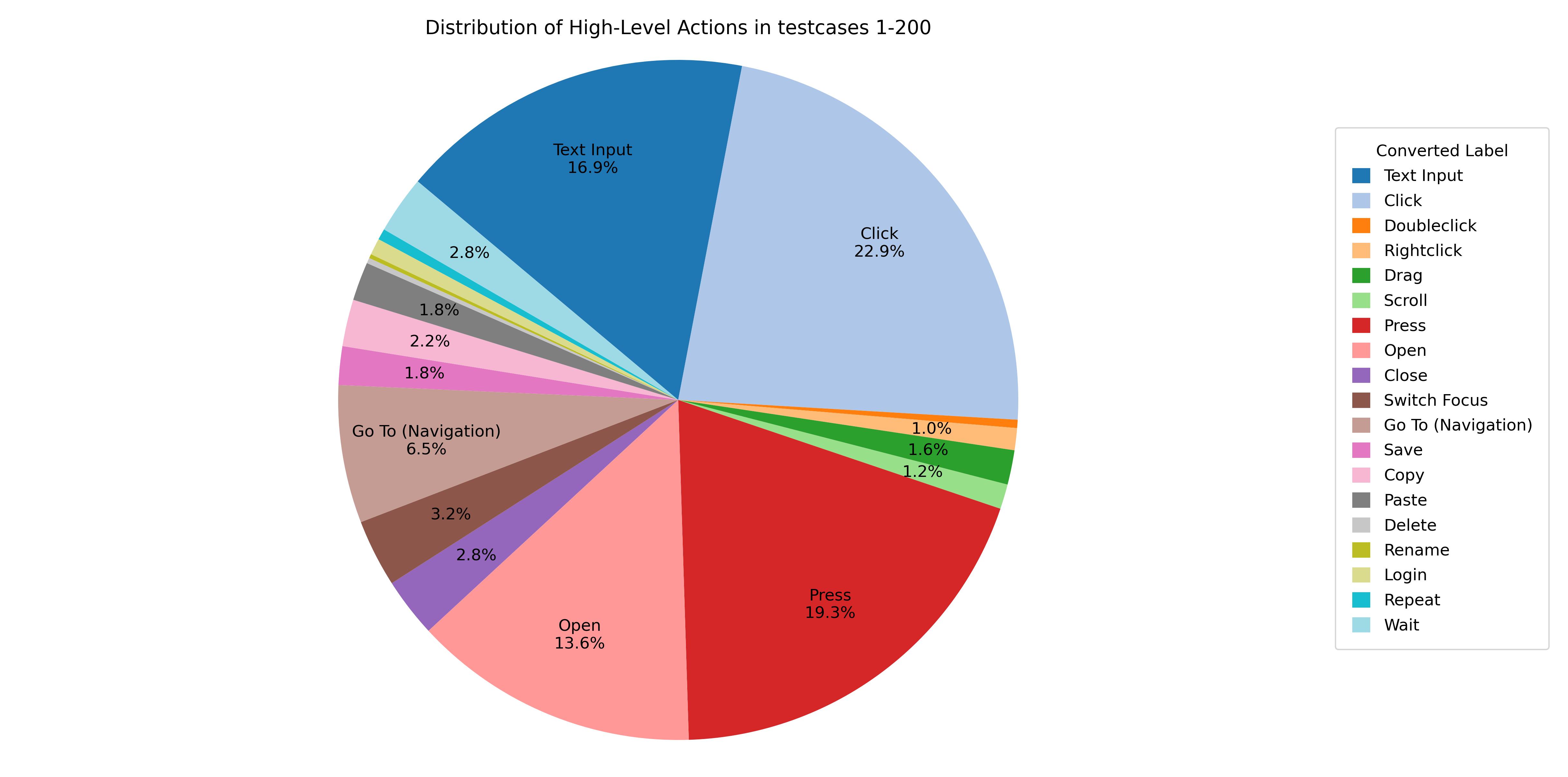}
    \caption{Distributions of 19 high level actions in testcases 1-200.}
    \label{appdx:19 high-level actions}
\end{figure*}

\begin{figure*}
    \centering
    \includegraphics[width=\linewidth]{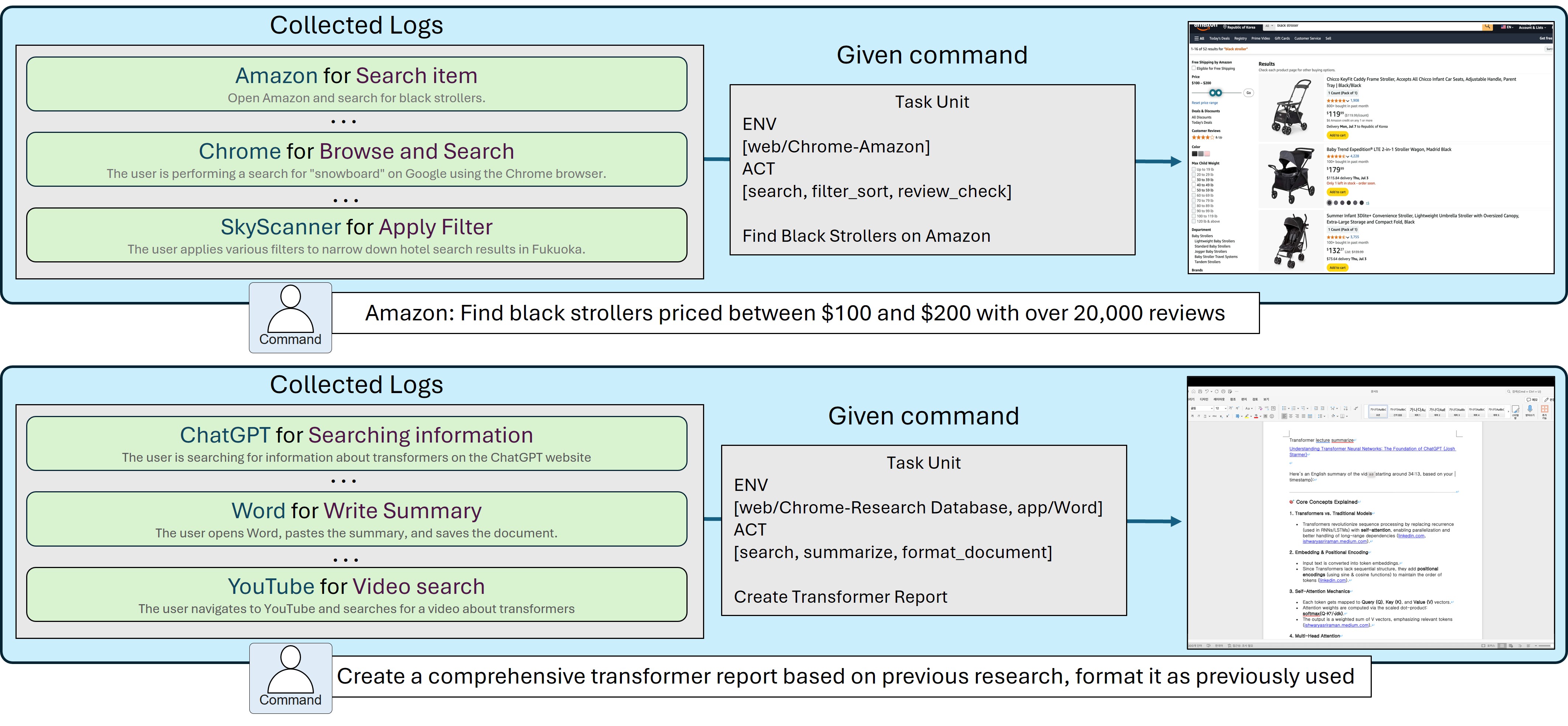}
    \caption{Examples of Task Retrieval process after user command input.}
    \label{appdx:loguse}
\end{figure*}

Figure `\ref{appdx:19 high-level actions} shows distributions of 19 high level actions performed while executing testcases.
Table \ref{appdx:task_details} shows 15 examples of 200 testcases.
Figure \ref{appdx:loguse} demonstrates detailed examples of the task retrieval process after user command input.

\begin{table*}[t]
\centering
\begin{tabular}{cl}
\toprule
\textbf{Category} & \textbf{Task Description} \\
\midrule
\multirow{5}{*}{\textbf{Log2Plan}} 
& 1. Create a comprehensive transformer report based on previous research, format it as previously used \\
& 2. Find transformer lecture video, summarize the content, and create a Word document in Documents folder \\
& 3. Extract highlighted sections about transformer and self-attention from research papers, create PowerPoint \\ &presentation slides in Documents folder \\
& 4.. Review next week schedule starting June 2nd, create weekly briefing document with upcoming meetings \\
& 5. Search for pet-friendly cafes using Chrome browser, create recommendation list based on location preferences \\
\midrule
\multirow{5}{*}{\textbf{Skyvern}} 
& 1. Locate cookie recipes with more than 1000 reviews and a rating of at least 4.5 stars \\
& 2. Locate non-stick oven-safe 10-piece cookware sets priced under \$150 \\
& 3. Find solutions on Apple's website if you forgot your Apple ID password \\
& 4. Find a Mexico hotel with deals for December 25--26 \\
& 5. Find a route from Miami to New Orleans, and provide the detailed route information \\
\midrule
\multirow{5}{*}{\textbf{Screen Agent Tasks}} 
& 1. Have a three-round conversation with ChatGPT about the Renaissance \\
& 2. Center the title of the presentation document \\
& 3. Check reviews of the movie \textit{Farewell My Concubine} on a movie review website \\
& 4. Select and add a smartphone to the shopping cart using the search bar on a shopping page \\
& 5. Download a paper on object detection from the TPAMI journal \\
\bottomrule
\end{tabular}
\caption{Examples of testcases.}
\label{appdx:task_details}
\end{table*}

\end{document}